%% file: acl_latex.tex
\pdfoutput=1

\documentclass[11pt]{article}

\usepackage[final]{acl}

\usepackage{times}
\usepackage{latexsym}
\usepackage{amsmath}
\usepackage[T1]{fontenc}
\usepackage[utf8]{inputenc}
\usepackage{microtype}
\usepackage{inconsolata}
\usepackage{graphicx}
\usepackage{enumitem}
\usepackage{xcolor}
\usepackage{subcaption}
\usepackage{fancyhdr}
\usepackage{tikz}
\usepackage{listings}
\usepackage{color}
\usepackage{booktabs}
\usepackage{multirow}
\usepackage{algorithmicx}
\usepackage{algorithm}
\usepackage{algpseudocode}
\usepackage{comment}
\usepackage{amssymb} 
\usepackage{tcolorbox}
\usepackage{epsfig}
\usepackage{boldline}
\usepackage{xspace}
\usepackage{hyperref}
\usepackage{url}
\usepackage{xcolor}     
\usepackage{amssymb}
\usepackage{pifont}
\usepackage{tikz}
\usepackage{color, colortbl}
\input{math_commands.tex}

\definecolor{Gray}{gray}{0.9}
\definecolor{myblue}{HTML}{d8ebf8}
\definecolor{COLOR_ZS}{HTML}{E6ECE3}

\def\*#1{\mathbf{#1}}
\def\name{\textsc{PromptRefine}\xspace}

\newtcolorbox{block}[1][]{
  colback=white, 
  colframe=black, 
  fonttitle=\bfseries,
  coltitle=black,
  boxrule=0.5pt,
  #1 
}

\newcommand{\cmark}{\ding{51}}%
\newcommand{\xmark}{\ding{55}}%

\title{\name: Enhancing Few-Shot Performance on Low-Resource Indic Languages with Example Selection from Related Example Banks} 

\author{Soumya Suvra Ghosal$^{1}$\thanks{Work done during internship at Adobe Research} \and Soumyabrata Pal$^{2}$ \and Koyel Mukherjee$^{2}$ \and  Dinesh Manocha$^{1}$\\
        \textsuperscript{1}University of Maryland, College Park;
        \textsuperscript{2}Adobe Research}

\begin{document}
\maketitle

\begin{abstract}
Large Language Models (LLMs) have recently demonstrated impressive few-shot learning capabilities through in-context learning (ICL). However, ICL performance is highly dependent on the choice of few-shot demonstrations, making the selection of the most optimal examples a persistent research challenge. This issue is further amplified in low-resource Indic languages, where the scarcity of ground-truth data complicates the selection process. In this work, we propose \name, a novel Alternating Minimization approach for example selection that improves ICL performance on low-resource Indic languages. \name leverages auxiliary example banks from related high-resource Indic languages and employs multi-task learning techniques to align language-specific retrievers, enabling effective cross-language retrieval. Additionally, we incorporate diversity in the selected examples to enhance generalization and reduce bias. Through comprehensive evaluations on four text generation tasks—Cross-Lingual Question Answering, Multilingual Question Answering, Machine Translation, and Cross-Lingual Summarization using state-of-the-art LLMs such as LLAMA-3.1-8B, LLAMA-2-7B, Qwen-2-7B, and Qwen-2.5-7B, we demonstrate that \name significantly outperforms existing frameworks for retrieving examples.\footnote{The code is available at \url{https://github.com/Soumya1612-Rasha/PromptRefine}.}
\end{abstract}

\input{sections/introduction}

\input{sections/related_works}
\input{sections/method}

\input{sections/experiments}

\input{sections/discussion}
\input{sections/conclusion}

\section{Limitations}
\label{app:limitations}

In Section~\ref{subsec:results}, we demonstrated the effectiveness of our proposed framework across diverse tasks. However, \name has the following limitations:
\begin{itemize}
    \item To improve text generation quality in low-resource Indic languages, \name depends on the availability of example banks from closely related, relatively high-resource Indic languages to provide additional context. We believe this is a reasonable assumption, as publicly available web-text resources are accessible for high-resource languages.
    
    \item As explained in Algorithm~\ref{algo:main}, our proposed approach involves three key steps: (1) identifying closely related example banks, (2) iterative refinement of retriever embeddings, and (3) diversity-based finetuning. These steps could have been arranged in several different configurations. However, we empirically tested several alternatives and found that our proposed approach consistently delivers the best performance improvements.
\end{itemize}

\bibliography{main}

\appendix
\input{sections/appendix}

\end{document}

%% file: math_commands.tex
\usepackage{amsmath,amsfonts,bm}

\def\eqref#1{equation~\ref{#1}}

\def\1{\bm{1}}

\DeclareMathAlphabet{\mathsfit}{\encodingdefault}{\sfdefault}{m}{sl}
\SetMathAlphabet{\mathsfit}{bold}{\encodingdefault}{\sfdefault}{bx}{n}

\DeclareMathOperator*{\argmax}{arg\,max}

%% file: sections/introduction.tex
\section{Introduction}
\label{sec:intro}

Large Language Models (LLMs) have recently made remarkable progress, demonstrating human-level performance across a wide range of tasks~\citep{adiwardana2020towards, wang2019superglue}. However, despite these advancements, most LLMs, such as LLaMA-3~\cite{dubey2024llama}, LLaMA-2~\cite{touvron2023llama}, and Qwen~\cite{yang2024qwen2}, are predominantly pre-trained on English texts, leading to significant performance disparities when applied to low-resource, non-English languages~\citep{ahuja2023megaverse}. The scarcity of ground-truth paired data in many low-resource languages makes text generation particularly challenging, as fine-tuning LLMs becomes infeasible in such settings. This issue is especially pronounced in lesser-known Indic languages such as Tibetan which has only around $5000$ Wikipedia articles compared to $6.6$M+ Wikipedia articles in English, despite having approximately $6$M speakers. In this work, we focus on downstream generation tasks with low-resource Indic languages, a critical challenge for making LLMs more widely accessible.

When downstream tasks have limited labeled data, few-shot learning or in-context learning (ICL) has emerged as a powerful and practical approach for text generation~\citep{tanwar2023multilingual, zhang2021differentiable, winata2021language, huang2023not, etxaniz2023multilingual}. ICL operates by providing the LLM with a prompt that consists of task-specific instructions and a set of input-output examples (demonstrations) to guide the model’s output generation for a specific input query. While ICL is computationally efficient (it requires no parameter updates), it faces two critical challenges when labeled data is scarce, particularly for low-resource Indic languages.

(\textbf{Relevance}) First, as in any learning task with limited data, the small size of the available example pool can result in a lack of relevant examples to guide the model effectively. In the context of ICL, this scarcity can severely degrade performance, as the quality of examples is crucial. Poor example selection can lead to performance worse than zero-shot scenarios, while optimal selection can achieve near state-of-the-art results~\citep{liu2021makes}. \textbf{(Diversity)} Second, existing example selection techniques, such as random selection or retrieving semantically similar examples~\cite{zhang2021differentiable, winata2021language, tanwar2023multilingual}, often struggle to account for diversity among the selected examples, which can further limit performance. An appropriate diverse set of selected examples in the prompt can improve generalization significantly.

\vspace{0.2cm}
\noindent\textbf{Proposed Approach.} In this work, we focus on an ICL approach for enhancing LLM performance in low-resource Indic languages. Specifically, our goal is to identify an optimal subset of guiding examples to be inserted into the prompt, without modifying the LLM parameters. Rather, we train retrievers to extract a near optimal set of examples (demonstrations) based on the input query. To this end, we propose a novel trainable approach \name that combines examples from auxiliary example banks associated with the related task. While \name is broadly applicable across different language families and even low-resource task clusters, in this study we focus on Indic languages~\cite{singh2024indicgenbench}, to enhance performance specifically on low-resource Indic languages through improved example selection. 

Given a specific input text generation task in a low-resource Indic language, we use the following set of key ideas to enhance model performance (see Algorithm \ref{algo:main}): 1) First, we utilize example banks from closely-related relatively high-resource \footnote{High resource language refers to a language in which a relatively large volume of pre-training data for training an LLM compiled from web-text is available.} Indic languages to provide relevant guidance, improving LLM performance on low-resource tasks. Therefore, the first step in our algorithmic approach given an input low-resource task is to select relevant high-resource languages (see Algorithm~\ref{algo:aux_select}) with associated example banks. 2) Second, merging these example banks is non-trivial, as each language-specific retriever is trained on a unique representation space. \textit{To address this, we adapt data-scarce multi-task learning techniques \citep{thekumparampil2021sample} to align the individual retrievers into a shared representation space, enabling cross-language retrieval.} 3) Third, we incorporate diversity in the selected examples to improve generalization and reduce bias during generation.

To empirically validate the effectiveness of \name, we evaluate its performance across four distinct text generation tasks on a subset of languages outlined in \citet{singh2024indicgenbench}: Cross-Lingual Question Answering (XorQA-In), Multilingual Question-Answering (XQuAD-In), Machine Translation (Flores-In), and Cross-Lingual Summarization (CrossSum-In), using recent LLMs such as LLAMA-3.1-8B~\citep{dubey2024llama}, LLAMA-2-7B~\citep{touvron2023llama}, Qwen-2-7B~\citep{yang2024qwen2}, and Qwen-2.5-7B~\citep{qwen2.5}. Our results demonstrate that \name significantly improves generation performance in all tasks compared to baseline approaches for example selection. Specifically, using LLAMA-3.1-8B as the LLM, \name achieves a Token-F1 improvement of $+\mathbf{16.07}$ and $+\mathbf{8.26}$ over zero-shot prompting and the current state-of-the-art retriever~\citep{ye2023compositional}, respectively, in the cross-lingual QA task. Similarly, using Qwen-2-7B, we observe an improvement in chrF1 of up to $+\mathbf{7.77}$ on the machine translation task compared to the baseline of selecting semantically similar examples. To ensure a comprehensive evaluation, we also test \name on proprietary LLMs such as GPT-3.5 and GPT-4~\cite{openai2024gpt4technicalreport}, where \name outperforms baseline retrievers on translation task, aligning with our previous findings.

%% file: sections/related_works.tex
\section{Related Works}
\label{sec:related_works}

\paragraph{In-Context Learning (ICL).} First introduced in \citet{brown2020language}, ICL has emerged as a powerful approach that enables large language models (LLMs) to ``learn by analogy'' by providing a few input-output examples as demonstrations, without requiring any update to model parameters. In recent years, a plethora of studies have provided insights on the underlying mechanism of ICL. \citet{saunshi2020mathematical} suggested that, by conditioning on a prompt, the task of predicting the next word becomes linearly separable, while \citet{xie2021explanation} observed that for ICL, the model infers a shared latent concept between the provided examples. A study pointed out that models do not rely as heavily on the provided input-output mappings as previously thought, indicating more nuanced learning dynamics in ICL~\citep{min2022rethinking}. \citet{chen2022improving, min2021metaicl, wei2023symbol} showed that the in-context learning ability of LLMs can be improved through self-supervised or supervised training. A group of studies have also explored to understand the factors affecting ICL~\citep{zhao2021calibrate, shin2022effect, wei2022emergent, yoo2022ground, wei2023larger} and the underlying working mechanism of ICL~\citep{olsson2022context, li2023transformers, pan2023context, dai2022can}. \citet{cahyawijaya2024llms} proposed query-alignment to improve the few-shot in-context learning performance of LLMs on low-resource languages.

\vspace{0.2cm}
\noindent\textbf{Example Selection for ICL.}  Despite the tremendous success, the performance of ICL is sensitive to specific settings, including the prompt template~\citep{wei2022chain, wang2022self, zhou2022large, zhang2022automatic}, the selection~\citep{liu2021makes, rubin2021learning, ye2023compositional, wang2024large} and order of demonstration examples~\citep{lu2021fantastically}, and
other factors~\citep{liu2024let}. Existing literature on example selection can be broadly categorized into two major groups: (1) Unsupervised methods depending on pre-defined metrics. \citet{liu2021makes} proposed selecting the closest neighbors as demonstrations, while \citet{levy2022diverse} observed that electing diverse demonstrations improves compositional generalization in ICL. \citet{wu2022self, nguyen2023context, li2023finding} explored using the output distributions of language models to select few-shot examples. (2) On the other hand, the other group of studies~\citep{li2023unified, luo2023dr, rubin2021learning, ye2023compositional} proposed fine-tuning a retriever model to select few-shot demonstrations. Recent works have also explored using reinforcement learning approaches~\citep{zhang2022active, scarlatos2023reticl} and Chain-of-thought reasoning~\citep{qin2023context} for example selection. In this study, we propose an alternating minimization approach for selecting the optimal set of in-context examples to enhance LLM performance on low-resource Indic languages.

%% file: sections/method.tex
\section{Preliminaries}
\label{sec:prelims}

\subsection{In-Context Learning} In-context learning (ICL) leverages the intrinsic abilities of language models to learn and infer new tasks without the need for parameter updates. Formally, let $\pi_{\text{LM}}$ be a language model with a vocabulary $\mathcal{V}$.
Consider a downstream generation task with input space $\mathcal{X}$ and output space $\mathcal{Y}$.
For a given test query $\*x_{\text{test}}\in \mathcal{X}$ and a retrieved subset of $K$ input-output pairs $\{(\*x_i,\*y_i)\}_{i=1}^{K}\in \mathcal{X}\times \mathcal{Y}$ describing the intended task, ICL generates the output $\*y_{\text{test}}\in \mathcal{Y}$ as follows:
\begin{equation*} \*y_{\text{test}} \sim \pi_{\text{LM}}(\cdot|\*x_{1}, \*y_{1}, \*x_{2}, \*y_{2}, \dots, \*x_{K}, \*y_{K}, \*x_{\text{test}}) \end{equation*}
Here, $\sim$ represents the sampling techniques commonly used in the literature, such as Greedy Sampling, Top-p Sampling~\citep{holtzman2019curious}, Top-k Sampling~\citep{fan2018hierarchical}, and Beam Search~\citep{freitag2017beam}. Each in-context example $a_i = (x_i, y_i)\in \mathcal{X}\times \mathcal{Y}$ is drawn from a training set example bank $\mathcal{D} = \left\{ (\*x_{i}, \*y_{i}) \right\}_{i=1}^{N}$ of input-output sequences.

\subsection{Example Retrieval for Few-shot learning}  
Our main goal is to train a retriever $\mathcal{R}_{\phi}(\*x_{\text{test}},\mathcal{D})$ model parameterized by $\phi$, that retrieves a set of in-context examples $\{a_i\}_{i=1}^K \subset \mathcal{D}$ given a test sample $\*x_{\text{test}}$ where typically $K \ll N$. Usually, $\phi:(\mathcal{X}\cup \mathcal{Y})^{\star}\rightarrow \mathbb{R}^d$ represents\footnote{$\mathcal{A}^{\star}$ denotes the smallest superset of $\mathcal{A}$ closed under set concatenation - known as Kleene operator. In this case $(\mathcal{X}\cup\mathcal{Y})^{\star}$ corresponds to the concatenation of strings in $\mathcal{X}$ and $\mathcal{Y}$.}  the embedding function that maps the text in the input space $\mathcal{X}$ into a $d$-dimensional vector representation. Such representations are subsequently
used to measure similarity between samples. Previous works have explored various retrieval strategies, ranging from random example selection to using off-the-shelf retrieval models~\citep{liu2021makes, wu2022self}, as well as fine-tuning the retriever's embeddings~\citep{ye2023compositional, rubin2021learning}.

\paragraph{Relevance Based Fine-tuning.} In this study, we leverage the framework proposed by \citet{rubin2021learning} to fine-tune an efficient dense retriever $\mathcal{R}_{\phi}$. 
The core idea is to train the retriever on a labeled dataset curated from the training data itself, optimizing it to select examples that serve as effective prompts. For each sample $(\*x, \*y) \in \mathcal{D}$, we generate a candidate set $\mathcal{A}=\{a_i\}_{i=1}^F$, where $a_i \in \mathcal{D}\setminus (\*x, \*y)$. The candidate set is selected using an unsupervised BM25 retriever: $\mathcal{A} = \text{BM25}((\*x, \*y), \mathcal{D})$ that simply retrieves $K$ examples with closest vector embedding to $\phi(\*x)$. Next, each candidate example $\*a_i \in \mathcal{A}$ is scored using a language model $\pi_{\text{Scorer}}$ based on its relevance to the sample $(\*x,\*y)$.

\begin{equation} s(\*a_i; (\*x, \*y)) = \pi_{\text{Scorer}}(\*y \mid \*a_i, \*x). \end{equation}
The best candidate example for the sample $(\*x, \*y)$ is selected as $\widetilde{\*a} = \argmax_{\*a_j} s(\*a_j; (\*x, \*y))$. Finally, the retriever $\mathcal{R}_{\phi}$ is fine-tuned to rank the candidate examples optimally (align with ranking induced by $\pi_{\text{Scorer}}$) by minimizing the negative log-likelihood under softmax loss:
\begin{align}
    &\min_{\phi} \mathcal{L}_{\text{rel}}(\mathcal{D}; \phi) = \frac{1}{N} \sum_{i=1}^N  \ell(\*x_i, \mathcal{A}_i) \\
    & \ell(\*x, \mathcal{A}; \phi) = -\text{log}\frac{e^{\text{sim}(\*x, \widetilde{\*a})}}{\sum_{\*a_i \in \mathcal{A}} e^{\text{sim}(\*x_i, \*a_i)}}
\end{align}
where $\text{sim}(\*a_i, \*a_j) = {\phi}(\*a_i)^{\top} \phi(\*a_j)$ measures the cosine similarity between the embeddings.
Note, that although a fine-tuned $\mathcal{R}_{\phi}$ is better aligned with the ranking induced by $\pi_{\text{Scorer}}$, it only optimizes for relevance and does not account for diversity. This often leads the finetuned retriever to choose near-identical samples thus hurting generalization.

\subsection{Determinantal Point Processes}
\label{subsec:dpp_prelims}
In this subsection, we introduce a recently studied framework Determinantal Point Processes (DPPs)~\citep{ye2023compositional} for example selection that ranks subsets of examples rather than individual ones. 
Introduced in \citet{macchi1975coincidence}, DPP's are elegant probabilistic models that have been extensively used in the literature~\citep{ye2023compositional, borodin2000distributions, benard1973detection, kulesza2012determinantal, liu2022determinantalpointprocesslikelihoods} to capture negative correlation among items.  

We leverage the DPP framework to promote diversity within the set of retrieved in-context examples. Formally, a point process $\mathcal{P}$ is called a DPP if, for any random subset $Y$ drawn according to $\mathcal{P}$, the probability that a subset $S$ is contained within $Y$ is given by:
\begin{equation*}
    \mathcal{P}(S \subseteq Y ) \propto \text{det}(\*Z_S)
\end{equation*}
where $\*Z \in \mathbb{R}^{n \times n}$ is a PSD similarity matrix, and $\*Z_S$ denotes the submatrix of $K$ corresponding to the rows and columns indexed by $S$. 
For in-context learning, given a test sample $\*x_{\text{test}}$, the input dependent similarity of any two examples $\*a_i,\*a_j$ is denoted by $\*Z_{ij}$ - formally, we model $\log \*Z_{ij}$ as 
\begin{equation*}
 \sum_{k\in \{i,j\}}\log\phi(\*a_k)^T\phi(\*x_{\text{test}})   + \log{\phi}(\*a_i)^{\top} \phi(\*a_j) 
\end{equation*}
where $\phi(\*a_i)^T\phi(\*x_{\text{test}}) \in \mathbb{R}^{+}$ measures the relevance of $\*a_i$ to input $\*x_{\text{test}}$ and $\phi(\*a_i)^T\phi(\*a_j)$ measures the similarity between the $i^{\text{th}}$ and $j^{\text{th}}$ example.

\section{Proposed Framework}
\label{sec:method}

\noindent\textbf{Problem Setup.} Despite the growing focus on evaluating the multilingual capabilities of large language models (LLMs), there remains a substantial performance gap between high-resource languages and those with limited web resources~\citep{ahuja2023megaverse, singh2024indicgenbench}. In this study, we aim to address this gap by enhancing the few-shot performance of low-resource Indic languages. To this end, we adopt the recently released IndicGen Benchmark~\citep{singh2024indicgenbench}, targeting a subset of Indic languages for our analysis. Specifically, we focus on the following low-resource languages:

\begin{block} 
\label{block:lang}
\textbf{Low/Mid-Resource Languages}: Bodo, Odia, Santali, Rajasthani, Manipuri, Awadhi, Marwari, and Maithili.

\vspace{0.1cm}
\textbf{Auxiliary High-Resource Languages}: Bengali, Hindi, Marathi, Gujarati, Kannada, Malayalam, Tamil, Telugu, Urdu. \end{block}

Additionally, we assume the availability of data from relatively high-resource Indic languages, which we refer to as the auxiliary dataset. This assumption is justified by the fact that these languages have relatively ample web-text resources~\citep{singh2024indicgenbench}.

\begin{algorithm}[t]
\caption{\name}
\label{algo:main}
\begin{algorithmic}[1]
\State \textbf{Input:} Low-resource language $\mathcal{T}$ example bank $\mathcal{D}^{\mathcal{T}}$; validation set $\mathcal{D}^{\mathcal{T}}_{\text{val}}$; Auxiliary example bank $\mathcal{D}^{\text{aux}}=\{\mathcal{D}^{\mathcal{H}_1}, \cdots, \mathcal{D}^{\mathcal{H}_M}\}$, number of iterations $\mathcal{I}$, Accuracy Metric Acc.

\State  $\alpha \gets \emptyset$
\State $\rho \gets \text{MBERT}$ 
\Comment{Initialize the retriever embedding with pre-trained Multi-lingual BERT encoder}

\For {iter in \{$1, \cdots, \mathcal{I}$\}} 
    \State $\Phi \gets \emptyset$
    \For {each dataset $\mathcal{D}^i \in \{\mathcal{D}^{\mathcal{T}}, \mathcal{D}^{\mathcal{H}_1}, \cdots, \mathcal{D}^{\mathcal{H}_M}\}$}
        \State $\phi_i \gets \min_{\rho} \mathcal{L}_{\text{rel}}(\mathcal{D}^i; \rho)$
        \State $\Phi \gets \Phi \cup \phi_i$
    \EndFor
    \State $\rho = \frac{1}{|\Phi|}\sum_{\theta\in\Phi} \theta$  \Comment{Merge the retriever embeddings}
     \State $\alpha_{\text{iter}} \gets \text{Acc}(\rho, \mathcal{D}^{\mathcal{T}}_{\text{val}})$ \Comment{Calculate validation accuracy}
    \State $\alpha \gets \alpha \cup \alpha_{\text{iter}}$ 
\EndFor
\State $\rho^* \gets \arg\max_{\rho} \alpha_{\text{iter}}$ 
\State $\Bar{\rho} \gets \min \mathcal{L}_{\text{DPP}}( \mathcal{D}^{\mathcal{T}} \cup \mathcal{D}^{\text{aux}}; \rho^*)$ 
\State \Return  $\Bar{\rho}$
\end{algorithmic}
\end{algorithm}

\subsection{Our Approach: \name}

To enhance the performance of language models on low-resource languages, we introduce \name, a three-step framework that: 1) identifies closely related high-resource Indic languages and leverages associated example banks (Section~\ref{subsubsec:aux_data_sel}), 2) iteratively refines retriever embeddings $\mathcal{R}_{\phi}$ (Section~\ref{subsubsec:iterative_prompt}), and 3) incorporates diversity-based finetuning of retriever $\mathcal{R}_{\phi}$ to rank subsets of in-context examples for a given input query (Section~\ref{subsubsec:dpp_finetune}). The complete approach is outlined in Algorithm~\ref{algo:main}. In what follows, we provide an in-depth overview of each component.

\subsubsection{Auxiliary Dataset Selection}
\label{subsubsec:aux_data_sel}

Despite the advent of numerous LLMs, low-resource Indic languages constitute only a negligible portion of their pre-training corpora, resulting in a suboptimal performance for generation tasks in these languages. To address this, we propose using relatively high-resource Indic languages, such as Hindi and Bengali, as auxiliary datasets. 
Our approach involves selecting auxiliary languages that are closely related to the target Indic low-resource language. Specifically, for each low-resource language, we compute the cosine similarity between the embeddings and those of the auxiliary languages. An auxiliary language is included if its similarity score surpasses a threshold parameter $\delta$,
as outlined in Algorithm~\ref{algo:aux_select} (Appendix~\ref{app:algorithm}). The underlying motivation for this approach is that for an input query in a low-resource Indic language, relevant examples that can guide the LLM in generation might not be present in the associated example bank - therefore,
by incorporating examples from the related high-resource languages, we provide the LLM additional context or rather relevant guidance that can help improve the model's performance on the input query. However, the following critical challenge is now raised:
\emph{"How can we effectively integrate the diverse information from auxiliary datasets for optimal performance?"}

\subsubsection{Iterative Prompt Refinement}
\label{subsubsec:iterative_prompt}
We denote the example bank of the low-resource target language $\mathcal{T}$ as $\mathcal{D}^{\mathcal{T}}=\left\{ (\*x_{i}, \*y_{i}) \right\}_{i=1}^{N}$, the selected set of auxiliary languages (using Alg. \ref{algo:aux_select}) as $\mathcal{H}=\{\mathcal{H}_1, \cdots, \mathcal{H}_M\}$ and the auxiliary example banks as $\mathcal{D}^{\text{aux}}=\{\mathcal{D}^{\mathcal{H}_1}, \cdots, \mathcal{D}^{\mathcal{H}_M}\}$. Note that the number of high-resource auxiliary example banks $M$ is determined by the threshold parameter $\delta$, as defined in Section~\ref{subsubsec:aux_data_sel}. Our goal is to train a single retriever $\mathcal{R}_\rho(\*x_{\text{test}},\mathcal{D}^{\text{aux}} \cup \mathcal{D}^{\mathcal{T}})$ - however the challenge is both the shared representation space for all example banks combined and the parameter weights $\phi$ are unknown. At the same time, the retriever must capture the specific traits of each individual language. Balancing these requirements requires an alternating optimization procedure.

To maximize the information gain from the auxiliary example banks, we propose an Alternating Minimization (AM) framework that alternately performs the following two steps successively until convergence 1) \textbf{(Specialize)} Fine-tune relevance-based retrievers $\{\mathcal{R}_{\phi_i}\}_{i}$ on each of several selected languages by retraining only on the example bank associated with the corresponding language (Step 7 in Alg. \ref{algo:main}). Intuitively, the goal of this step is to allow a particular retriever to acquire task / language-specific knowledge associated with the corresponding language.
Note that each of the individual retrievers $\{\mathcal{R}_{\phi_i}\}_{i}$ is initialized with pre-trained multilingual BERT encoder weights at the beginning of first iteration - in subsequent iterations, all the individual retrievers are initialized with shared parameter weights $\rho$ computed in the next step 2) \textbf{(Merge)} merges the individual retrievers $\{\mathcal{R}_{\phi_i}\}_{i}$ into a single retriever $\mathcal{R}_{\rho}$ by simple parameter averaging to obtain a \textit{shared representation space enabling cross-language retrieval} (Step 10 in Alg. \ref{algo:main}). Intuitively, the shared retriever $\mathcal{R}_{\rho}$ encapsulates the diverse knowledge learnt by each individual retriever.

In essence, our alternating minimization algorithm (Alg. \ref{algo:main}) alternately finetunes the individual retriever on language-specific example bank and creates a merged retriever enabling a shared representation space for $\mathcal{I}$ iterations.
We denote the model with the highest validation accuracy on the target language $\mathcal{T}$ after $\mathcal{I}$ iterations as $\rho^*$ (Step 11 in Alg. \ref{algo:main}).

\input{tables/xorqa}
\input{tables/flores_xx_en}

\subsubsection{Divsersity-induced finetuning}
\label{subsubsec:dpp_finetune}

A major limitation of relevance-based finetuning is that the in-context examples are retrieved solely based on relevance, thereby ignoring diversity and inter-relationship among the selected examples~\citep{ye2023compositional}. To overcome this challenge, we leverage the DPP framework to enhance diversity within the retrieved in-context examples. Specifically, we obtain the final retriever model by fine-tuning $\rho^*$ on the merged dataset $\widetilde{\mathcal{D}} = \mathcal{D}^{\mathcal{T}} \cup \mathcal{D}^{\text{aux}}$. Due to the technically involved training procedure of the DPP framework via contrastive learning, we delegate the training details to the Appendix~\ref{app:dpp}.
Note that during inference, we use the retriever model further fine-tuned in the DPP framework to retrieve both diverse and relevant samples~\citep{ye2023compositional}. Specifically, we perform MAP inference using the model within the DPP framework - however exact MAP inference is NP-Hard~\citep{ko1995exact} since it involves an exact search over an exponentially large collection of subsets of examples. However, in our setting, we can use a greedy but computationally efficient procedure to build the subset by inserting one example at a time~\citep{chen2018fastgreedymapinference}.

%% file: tables/xorqa.tex
\newcolumntype{?}{!{\vrule width 1pt}}
\newcolumntype{a}{>{\columncolor{myblue}}c}
\begin{table*}[!t]
      \centering
       
        \resizebox{\textwidth}{!}{%
        \begin{tabular}{cccccc?ccc?ccc}
        \toprule
       \multirow{2}{1.75cm}{\centering Aux. Data Used} & \multirow{2}{1.75cm}{\centering Finetuning-Based} & \multirow{2}{*}{Methods} & \multicolumn{3}{c?}{Bodo} & \multicolumn{3}{c?}{Manipuri} & \multicolumn{3}{c}{Maithili} \\ 
       \cmidrule{4-12}

       & & & QW-2-7B & QW-2.5-7B & LM-3.1-8B & QW-2-7B & QW-2.5-7B & LM-3.1-8B & QW-2-7B & QW-2.5-7B & LM-3.1-8B  \\

       \midrule
     
       \xmark & \xmark & Zero-shot & 0.68 & 0.21 & 1.20 & 0.68  & 0.44 & 0.97 & 3.41 & 2.72 & 13.05 \\
       \xmark & \xmark & Random &  4.57 &  4.02 & 5.83 & 4.72 & 3.80 & 6.10 & 13.97 & 12.02 & 18.73 \\
       \xmark & \xmark & BM25 & 6.20 & 6.47 & 9.41 & 5.68 & 4.14 & 6.98 & 13.21 & 11.77 & 20.15  \\
       \xmark & \xmark & Top-K & 5.90 & 6.37 & 5.10 & 5.80 & 4.21 & 6.08 & 13.25 & 12.95 & 18.02  \\
       \xmark & \xmark & Diverse &  5.88 & 5.92 & 5.34 & 5.54 & 4.09 & 5.98 & 13.27 & 12.53 & 18.29\\
       \xmark & \cmark & EPR & 6.29 & 7.15 & 7.81 & 6.40 & 4.91 & 7.22 & 14.85 & 14.03 & 20.18  \\
       \xmark & \cmark & CEIL & 7.83 & 7.33 & 8.20 & 6.73 & 6.17 & 9.33 & 16.80 & 15.72 & 21.96\\
       \midrule
       \cmark & \cmark & EPR &  6.61 & 7.99 & 7.98 & 6.34 & 4.58 & 7.18 & 14.27 & 13.70 & 19.71 \\
       \cmark & \cmark & CEIL & 7.95 & 8.07 & 9.01 & 6.62 & 6.09 & 9.28 & 16.84 & 15.49 & 22.38\\
     \rowcolor{myblue} \cmark & \cmark & \name~(Ours) & \textbf{13.68} & \textbf{11.40} & \textbf{17.27} & \textbf{12.79} & \textbf{11.50} & \textbf{19.54} & \textbf{24.56} & \textbf{23.37} & \textbf{25.59} \\
    
     & & Absolute Gain $(\Delta)$ & +6.05 & +3.33 & +8.26 & +6.17 & +5.33 & +10.21 & +7.76 & +7.65 & +3.21 \\

      \bottomrule
    \end{tabular}%
        }
 \caption{\small \textbf{Evaluation on XorQA-In.} We report Token-F1 results for performance on Bodo, Manipuri and Maithili. The evaluation includes three LLMs: Qwen-2-7B (QW-2-7B), Qwen-2.5-7B (QW-2.5-7B), and LLAMA-3.1-8B (LM-3.1-8B).}
\label{tab:xorqa}
\end{table*}

%% file: tables/flores_xx_en.tex
\newcolumntype{?}{!{\vrule width 1pt}}
\newcolumntype{a}{>{\columncolor{myblue}}c}
\begin{table*}[!t]
      \centering
        \resizebox{\textwidth}{!}{%
        \begin{tabular}{cccccc?ccc?ccc}
        \toprule
       \multirow{2}{1.75cm}{\centering Aux. Data Used} & \multirow{2}{1.75cm}{\centering Finetuning-based} & \multirow{2}{*}{Methods} & \multicolumn{3}{c?}{Santali $\rightarrow$ English} & \multicolumn{3}{c?}{Rajasthani $\rightarrow$ English} & \multicolumn{3}{c}{Manipuri $\rightarrow$ English} \\ 
       \cmidrule{4-12}
       & & & LM-2-7B & LM-3.1-8B & QW-2-7B & LM-2-7B & LM-3.1-8B & QW-2-7B & LM-2-7B & LM-3.1-8B & QW-2-7B  \\
       \midrule
       \xmark & \xmark & Zero-shot & 5.89 & 16.83 & 9.95 & 21.94 & 37.90 & 34.09 & 10.62 & 14.40 & 10.04  \\
       \xmark &  \xmark & Random & 8.54 & 16.92 & 10.85 & 22.08 & 37.69 & 37.45 & 11.96 & 16.57 & 11.87  \\
       \xmark & \xmark & BM25 & 9.77 & 17.55 & 11.60 & 24.35 & 36.62 & 36.92 & 11.67 &  17.81 & 12.18 \\
       \xmark & \xmark & Top-K & 9.98 & 18.66 & 11.98 & 25.12 & 38.33 & 37.05 & 11.81 & 17.40 & 14.93 \\
       \xmark &  \xmark & Diverse & 8.86 & 18.67 & 11.64 & 24.88 & 39.12 & 38.13 & 12.44 & 18.13 & 13.82 \\
       \xmark & \cmark & EPR & 10.35 & 18.59 & 12.09 & 25.99 & 40.09 & 37.97 & 13.90 & 18.99 & 18.79 \\
       \xmark & \cmark & CEIL & 10.68 & 18.41 & 12.20  & 26.04 & 41.85 & 38.77 & 14.02 & 19.53 & 19.47\\
       \midrule
       \cmark & \cmark & EPR & 10.27 & 18.73 & 12.38 & 25.77 & 41.60 & 38.30 & 13.88 & 18.82 & 20.18 \\
       \cmark & \cmark & CEIL & 10.63 & 17.90 & 13.01 & 25.72 & 41.93 & 40.04 & 14.15 & 18.97 & 20.23 \\
     \rowcolor{myblue} \cmark & \cmark & \name~Ours) & \textbf{15.14} & \textbf{23.58} & \textbf{15.48} & \textbf{31.75} & \textbf{45.88} & \textbf{43.43} & \textbf{18.69} & \textbf{23.90} & \textbf{22.70} \\
       &  & Absolute Gain ($\Delta$) & +4.48 & +4.85 & +2.47 & +5.71 & +3.95 & +3.39  & +4.54 & +4.37 & +3.27\\
      \bottomrule
    \end{tabular}%
        }
 \caption{\small \textbf{Evaluation on Flores-In.} We report chrF1 results for translation performance from three low-resource languages to English. The evaluation includes three LLMs: LLAMA-2-7B (LM-2-7B), LLAMA-3.1-8B (LM-3.1-8B), and Qwen-2-7B (QW-2-7B).}
\label{tab:flores}
\vspace{-0.5cm}
\end{table*}

%% file: sections/experiments.tex
\input{tables/cross_sum}
\section{Experiments}
\label{sec:exp}

\paragraph{Implementation Details.} In this study, we evaluate the multilingual and cross-lingual language generation capabilities of various LLMs on low-resource Indic languages mentioned in Section~\ref{sec:method}. Our experiments involve a range of recently released open-source LLMs, including Qwen-2-7B~\citep{yang2024qwen2}, Qwen-2.5-7B~\citep{qwen2.5}, LLAMA-2-7B~\citep{touvron2023llama}, and LLAMA-3.1-8B~\citep{dubey2024llama}, as well as proprietary models such as GPT-3.5 and GPT-4~\citep{openai2024gpt4technicalreport}. For open-source LLMs, we use the same model for scoring ($\pi_{\text{Scorer}}$) and inference ($\pi_{\text{LM}}$). For all experiments, we set the number of in-context examples
as $K=16$. Following previous literature~\citep{ye2023compositional, rubin2021learning}, we sort the in-context examples in ascending order of their similarity to the input text.

In our proposed approach, \name, the retriever embeddings are initialized using a pre-trained multilingual BERT encoder\footnote{google-bert/bert-base-multilingual-cased}. For fine-tuning, we use an Adam optimizer with a batch size of $64$ and a learning rate of $1e-4$. For relevance-based training, we fine-tune the model for $\mathcal{I}=10$ iteration with $120$ epochs in each iteration. We further fine-tune for $10$ epochs during DPP-based training. All experiments are run using the configurations listed in Appendix~\ref{app:hardware}.

\vspace{0.1cm}
\noindent\textbf{Baselines.} In this work, we introduce \name, a fine-tuning-based retrieval framework designed to enhance in-context example selection for low-resource Indic languages. For a comprehensive evaluation, we compare \name with fine-tuning-free retrievers such as Random, BM25~\cite{robertson2009probabilistic}, Top-K, and Diverse. Among fine-tuning-based retrievers, we benchmark against EPR~\citep{rubin2021learning} and CEIL~\cite{ye2023compositional}. We refer the reader to
Appendix~\ref{app:baseline} for a detailed description of
the baselines used in this study.

\vspace{0.1cm}
\noindent\textbf{Datasets.} We evaluate \name on four diverse tasks, as outlined in \citet{singh2024indicgenbench}, to comprehensively assess its retrieval capabilities for low-resource Indic languages. These tasks include Cross-Lingual Question Answering (XorQA-In-XX), Multilingual Question Answering (XQuAD-IN), Machine Translation (Flores-In-XX-En), and Cross-Lingual Summarization (CrossSum-IN). We refer the reader to
Appendix~\ref{app:datasets} for a detailed description of
the datasets used in this study. We present the prompt template used for different datasets in Table~\ref{tab:prompts} (Appendix).

\vspace{0.1cm}
\noindent\textbf{Evaluation Metrics.} Following prior work~\citep{singh2024indicgenbench}, we report the Character-F1 (chrF1) metric~\citep{popovic2015chrf, gala2023indictrans2highqualityaccessiblemachine} for cross-lingual summarization and translation tasks, as token-level metrics like ROUGE and BLEU are considered unreliable for low-resource languages~\citep{bapna2022building, singh2024indicgenbench}. For QA tasks (XQuAD-IN and XorQA-IN), we use the Token-F1 metric to maintain consistency with the existing literature~\citep{singh2024indicgenbench}.

\begin{figure}[!t]
    \centering
    \includegraphics[width=0.85\columnwidth]{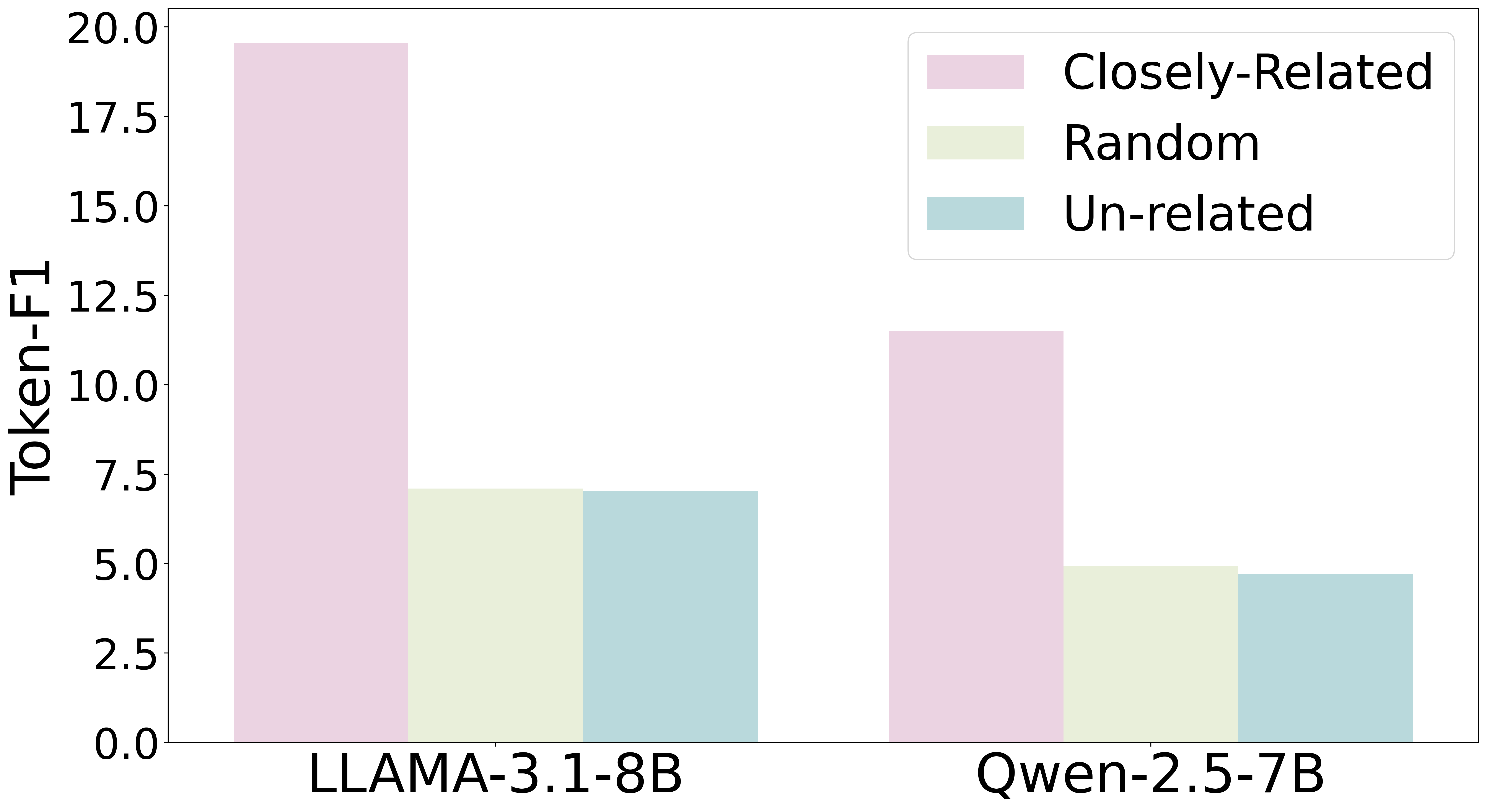}
    \caption{
    \small Token-F1 evaluation on the cross-lingual QA task in Manipuri, with retrievers trained using different auxiliary high-resource example banks: (1) Closely related language, (2) Random language, and (3) Unrelated language.}
    \label{fig:ablation_close_lang}
    \vspace{-0.3cm}
\end{figure}

\subsection{Main Results}
\label{subsec:results}

\noindent\textbf{Evaluation on open-source LLMs.} We present the evaluation results on state-of-art open-source models, including LLAMA-3.1-8B, LLAMA-2-7B, Qwen-2-7B, and Qwen-2.5-7B, for Cross-Lingual QA task (Table~\ref{tab:xorqa}), Machine Translation task (Table~\ref{tab:flores}), Cross-lingual Summarization task (Table~\ref{tab:cross_sum}), and Multi-lingual QA task (Table~\ref{tab:xquad}). Our empirical findings yield several key insights: (1) Across all diverse tasks, \name consistently outperforms both finetuning-based and finetuning-free baselines, with a improvement of {up to 2.09x} compared to current state-of-art. Notably, for Cross-lingual QA task, for Manipuri, \name improves the Token-F1 by $+10.21$ over CEIL~\citep{ye2023compositional}. (2) Compared to fine-tuning free baselines such as Top-K, \name provides a staggering improvement {of upto 3.38x} on low-resource language such as Bodo. Further, when compared to finetuning-based retrievers like EPR~\cite{rubin2021learning} and CEIL, \name consistently delivers superior performance across all tasks, as exemplified by a $+3.21$ Token-F1 improvement on Maithili. (3) Interestingly, even when auxiliary datasets are available, other finetuning-based retrievers such as EPR~\cite{rubin2021learning} and CEIL~\cite{ye2023compositional} do not exhibit significant improvements. This finding highlights the importance of learning a better representation space to effectively integrate and leverage auxiliary data. In contrast, \name successfully utilizes auxiliary data to generate a richer set of in-context examples, as reflected by its substantial performance gains across multiple tasks.

\vspace{0.1cm}
\noindent\textbf{Evaluation on proprietary LLMs.} To ensure a comprehensive evaluation, we also test \name on proprietary LLMs such as GPT-3.5/4~\cite{openai2024gpt4technicalreport}. However, since we do not have access to the output logits in proprietary models, required for scoring candidate examples, we use a different open-source LLM architecture for scoring. To be specific, for this experiment, we employ LLAMA-3.1-8B as the scorer model $\pi_{\text{Scorer}}$. The results for the translation task are reported in Table~\ref{tab:flores_close} 
(Appendix). Consistent with our findings using open-source LLMs, \name significantly improves generation quality, as measured by chrF1, compared to other retrieval methods.

%% file: tables/cross_sum.tex
\newcolumntype{?}{!{\vrule width 1pt}}
\newcolumntype{a}{>{\columncolor{myblue}}c}
\begin{table*}[!t]
      \centering
       
        \resizebox{\textwidth}{!}{%
        \begin{tabular}{ccccc?cc?cc?cc}
        \toprule
       \multirow{2}{1.75cm}{\centering Aux. Data Used} & \multirow{2}{1.75cm}{\centering Finetuning-Based} & \multirow{2}{*}{Methods} & \multicolumn{2}{c?}{Rajasthani} & \multicolumn{2}{c?}{Manipuri} & \multicolumn{2}{c?}{Marwari} & \multicolumn{2}{c}{Awadhi} \\ 
       \cmidrule{4-11}

       & & & LM-3.1-8B & QW-2-7B & LM-3.1-8B & QW-2-7B & LM-3.1-8B & QW-2-7B &  LM-3.1-8B & QW-2-7B  \\
       \midrule
       \xmark & \xmark & Zero-shot  &  1.10 & 3.06 & 0.15 & 0.36 & 0.16 & 2.09 & 0.19 & 2.16\\
       \xmark & \xmark & Random  &  8.38 & 5.93 & 3.14 & 3.88 & 7.38 & 6.05 & 9.10 & 4.20 \\
      \xmark & \xmark & BM25  &  7.86 & 5.85 & 3.52 & 3.92 & 7.81 & 5.54 & 8.40 & 4.83 \\
      \xmark & \xmark & Top-K & 11.03 & 6.08 & 4.43 & 4.50 & 8.91 & 5.91 & 9.36 & 6.07\\
       \xmark & \xmark & Diverse & 10.77 & 6.21 & 4.39 & 4.68 & 8.93 & 5.78 & 9.05 & 6.11\\
       \xmark & \cmark & EPR &  11.25 & 7.16 & 5.11 & 5.13 & 9.33 & 6.32 & 10.59 & 6.44\\
       \xmark & \cmark & CEIL & 11.93 & 7.39 & 5.49 & 5.92 & 10.49 & 6.70 & 11.92 & 7.02\\
       \midrule
       \cmark & \cmark & EPR & 11.22 & 7.09 & 5.08 & 5.59 & 9.11 & 6.25 & 11.36 & 6.58 \\
        \cmark & \cmark & CEIL & 11.50 & 7.28 & 5.71 & 6.04 & 10.20 & 6.67 & 12.41 & 7.33\\
     \rowcolor{myblue} \cmark & \cmark & \name~(Ours) & \textbf{15.88} & \textbf{9.43} & \textbf{6.67} & \textbf{6.92} & \textbf{15.95} & \textbf{8.06} & \textbf{15.36} & \textbf{8.49}\\
     & & Absolute Gain $(\Delta)$ & +3.95 & +2.04 & +0.96 & +0.88 & +5.46 & +1.36 & +2.95 & +1.47 \\
       
      \bottomrule
    \end{tabular}%
        }
 \caption{\small \textbf{Evaluation on CrossSum-In.} We report chrF1 results for the summarization performance of an English text into four low-resource languages. The evaluation includes two recent LLMs: LLAMA-3.1-8B (LM-3.1-8B), and Qwen-2-7B (QW-2-7B). }
\label{tab:cross_sum}
\end{table*}

%% file: sections/discussion.tex
\section{Discussion}
\label{sec:discussion}
Note, ablation studies on threshold parameter $\delta$ for selecting related languages and number of ICL examples are deferred to Appendix \ref{app:discussion}.

\begin{figure}[!t]
    \centering
    \includegraphics[width=0.8\columnwidth]{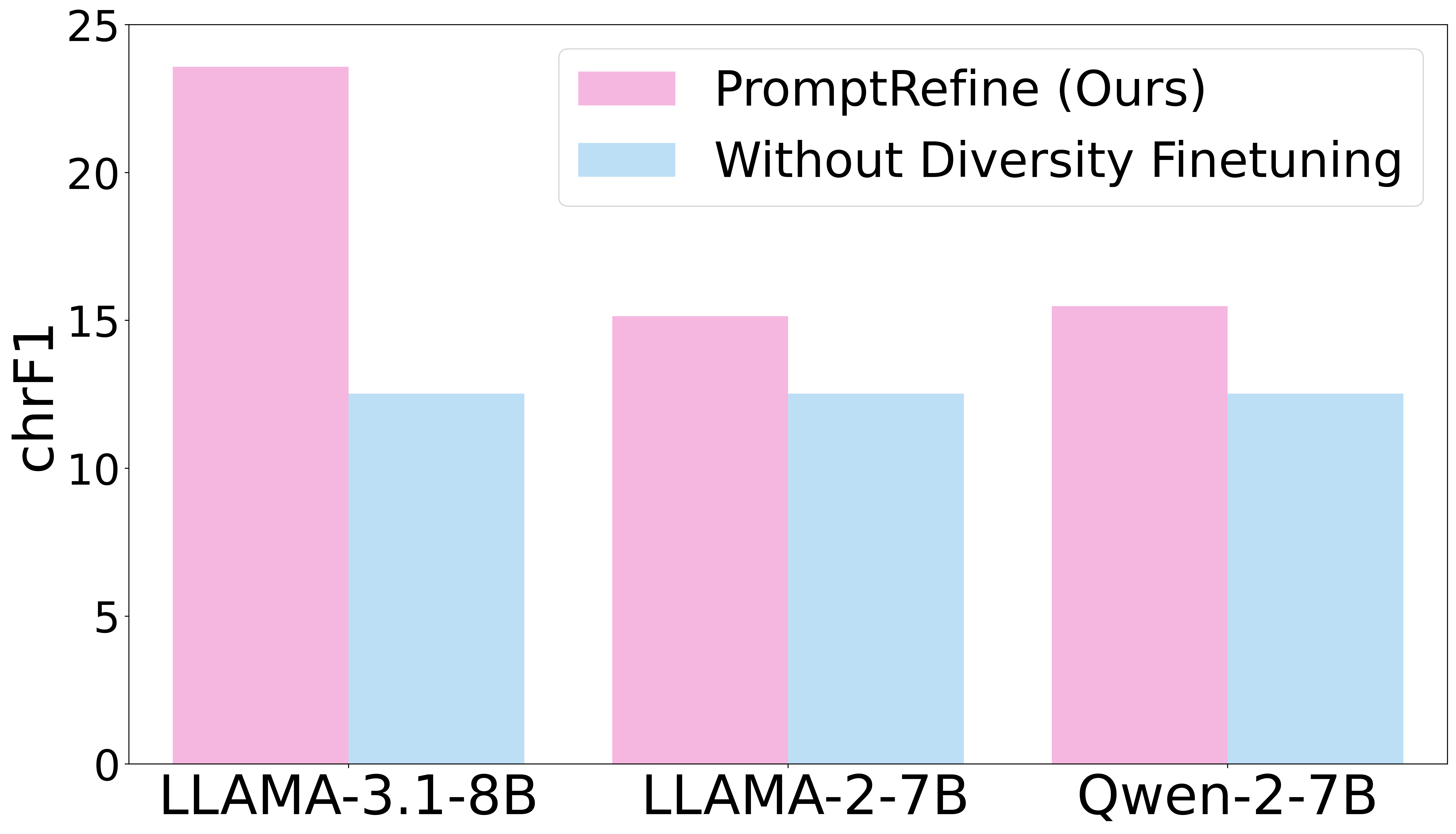}
    \caption{
    \small
    To highlight the importance of DPP training, we compare the quality of responses generated with and without diversity-induced fine-tuning on the task of translating from Santali to English.}
    \label{fig:ablations_dpp}
    \vspace{-0.6cm}
\end{figure}

\noindent\textbf{Why is choosing a closely related auxiliary example bank important?} In Section~\ref{subsubsec:aux_data_sel}, we discussed the approach of selecting a closely related high-resource Indic language as an auxiliary example bank for a target low-resource language. To highlight importance of selecting the right set of related languages, we present ablation studies in this section.

We evaluate three setups: (1) Our method (Alg. \ref{algo:aux_select}) of selecting the most closely related language as the auxiliary example bank, (2) selecting a random language, and (3) selecting the most unrelated language. We show the few-shot performance on cross-lingual QA task in Figure~\ref{fig:ablation_close_lang}, with retrievers trained under each setup. Our findings show that: (1) As expected, using the most closely related language as the auxiliary example bank yields the best performance, as it provides relevant guidance to the LLM, and (2) Selecting a random or unrelated language results in little to no improvement, with performance remaining close to the relevance-based EPR baseline~\cite{rubin2021learning} that only uses target-language specific data.

\vspace{0.1cm} 
\noindent\textbf{Importance of Diversity-Induced Fine-Tuning.}
A key factor contributing to the improved performance of \name is the retrieval of a diverse set of in-context examples. Merging auxiliary example banks imply a larger overall example bank and therefore incorporating diversity during example selection becomes important to improve generalization. In this section, we empirically demonstrate the significance of diversity-induced fine-tuning (Section~\ref{subsubsec:dpp_finetune}). 
Figure~\ref{fig:ablations_dpp} compares the text generation quality in Santali-to-English translation task, where examples are selected using two retriever models namely our approach \name with and with-out diversity-induced fine-tuning - in the latter, we skip line 15 in Alg. \ref{algo:main}. Results clearly show that incorporating diversity training leads to performance improvements across different LLMs, underscoring its effectiveness.

\begin{figure}[!t]
    \centering
    \includegraphics[width=0.88\columnwidth]{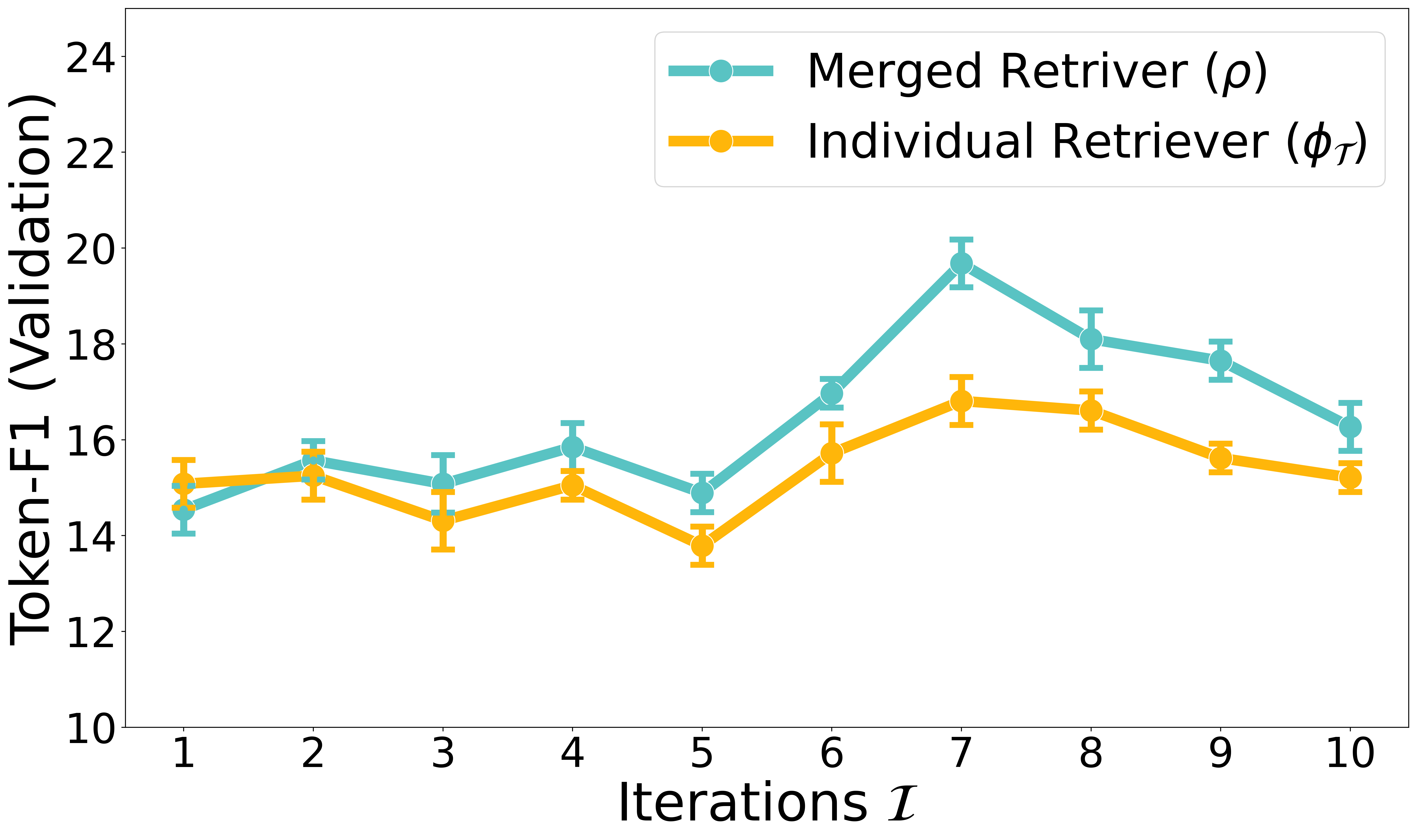}
    \caption{
    \small We plot the validation accuracy (measured by Token-F1) obtained by individual and merged retrievers over different iterations of finetuning.}
    \label{fig:ablations_am}
    \vspace{-0.6cm}
\end{figure}

\vspace{0.1cm}
\noindent\textbf{Importance of Alternating-Minimization approach.} In Section~\ref{subsubsec:iterative_prompt}, we proposed an alternating minimization approach to alternately fine-tune the retriever on language-specific data and merge the parameters in order to learn a shared representation space that captures the knowledge associated with both the target language $\mathcal{T}$ and the auxiliary languages $\mathcal{H}$. In Figure~\ref{fig:ablations_am}, we plot the validation accuracy across different iterations for the merged retriever $\rho$ and the target language retriever $\phi_{\mathcal{T}}$. The results show that merging the retriever embeddings provides a significant boost in validation accuracy over the iterations, thereby justifying the effectiveness of the parameter-averaging strategy.

%% file: sections/conclusion.tex
\section{Conlusion}
\label{sec:conclusion}

The pre-training data for state-of-the-art LLMs~\cite{dubey2024llama, yang2024qwen2} predominantly comprises English text, resulting in suboptimal performance on low-resource Indic languages~\cite{singh2024indicgenbench}. To address this, we propose a novel Alternating Minimization approach \name for example selection that improves ICL performance on low-resource Indic languages. Our approach follows a three-step framework: (1) identifying closely related high-resource Indic languages and utilizing their example banks, (2) iteratively refining retriever embeddings, and (3) employing diversity-based finetuning to rank subsets of in-context examples for a given input query. Comprehensive testing across various tasks and LLMs demonstrates that \name significantly enhances text generation quality on low-resource Indic languages.

%% file: sections/appendix.tex
\section{Software and Hardware}
\label{app:hardware}
We run all experiments with Python 3.12.4, PyTorch 2.2.0, and Transformers 4.43.3. For all experimentation, we use four Nvidia RTX A6000 GPUs.

\section{Extended Results}
\label{app:results}

In the main paper, we showed the effectiveness of \name for cross-lingual QA (Table~\ref{tab:xorqa}), machine translation (Table~\ref{tab:flores}), and summarization tasks (Table~\ref{tab:cross_sum}). In Table~\ref{tab:xquad}, we further present results on multi-lingual QA task for Odia language. Note that Xquad-In~\citep{singh2024indicgenbench} only includes medium-resource languages; therefore, we report results for only one language for this task. For a comprehensive evaluation, we also test \name on proprietary LLMs such as GPT-3.5/4~\cite{openai2024gpt4technicalreport}. We report results in Table~\ref{tab:flores_close}.

\section{Extended Discussion}
\label{app:discussion}

\begin{figure}[!t]
    \centering
    \includegraphics[width=\columnwidth]{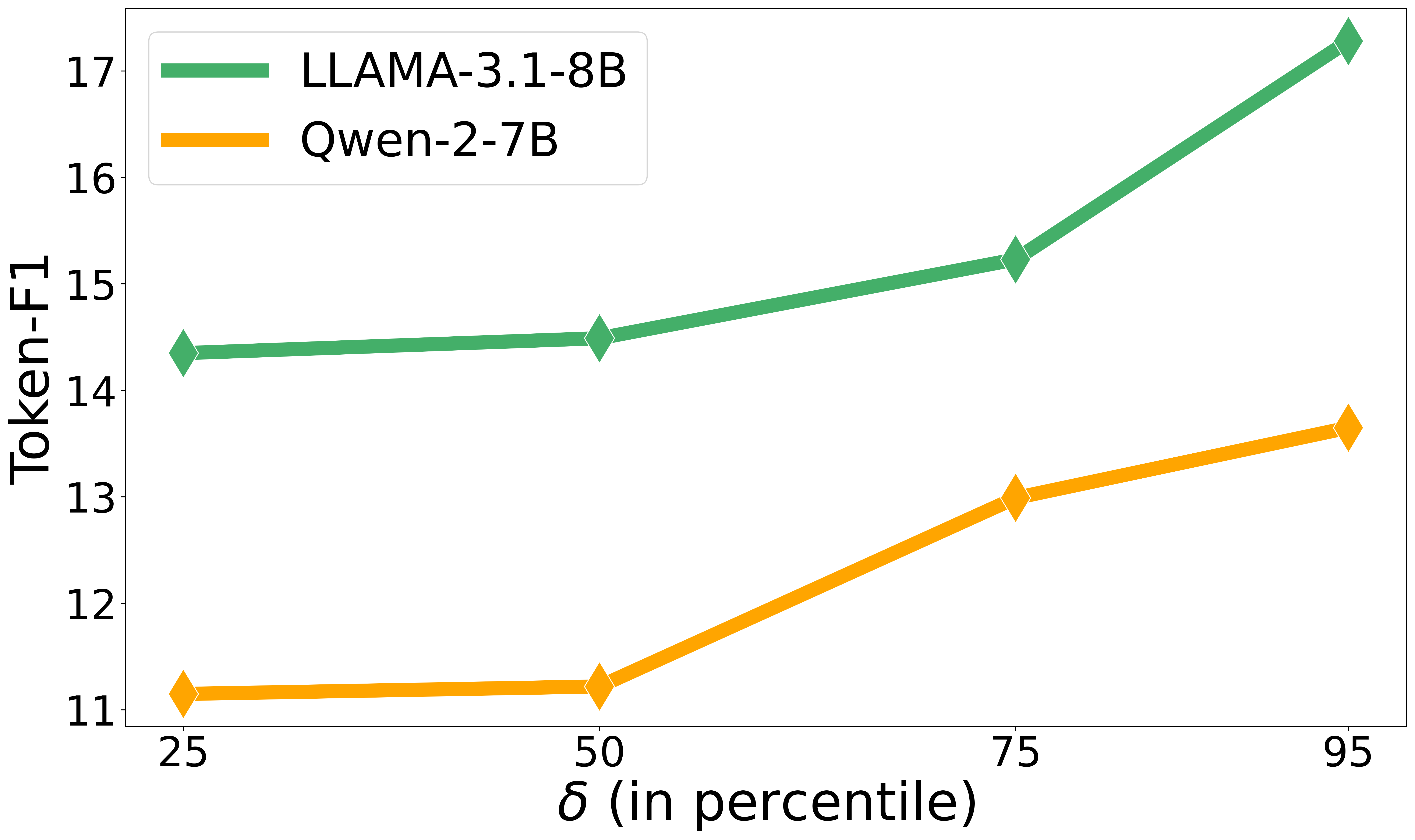}
    \caption{
    \small We vary the threshold parameter $\delta$ to evaluate its impact on model performance. Our results show that setting $\delta$ to the 95th percentile of cosine similarity values between target language embeddings and high-resource languages yields optimal performance. The evaluation task for this experiment is cross-lingual QA on Bodo.}
    \label{fig:ablations_threshold}
\end{figure}

\noindent\textbf{Ablation Study on Threshold Parameter $\delta$.} In this section, we present an ablation study on the threshold parameter $\delta$, which governs the selection of high-resource languages in the auxiliary dataset. A lower $\delta$ expands the inclusion to a broader set of high-resource languages, potentially introducing less relevant examples. On the other hand, a higher $\delta$ is more selective, which may restrict the diversity of examples and reduce the contextual richness. Our goal is to identify the optimal $\delta$ that balances these factors, ensuring the auxiliary dataset consists of examples from closely related high-resource languages that effectively improve LLM generation performance. Figure~\ref{fig:ablations_threshold} illustrates the translation performance from three different low-resource languages to English as a function of varying $\delta$. The results suggest that setting $\delta$ to the 95th percentile of the cosine similarity values between target language embeddings and high-resource languages yields optimal performance.

\begin{figure}[!t]
    \centering
    \includegraphics[width=\columnwidth]{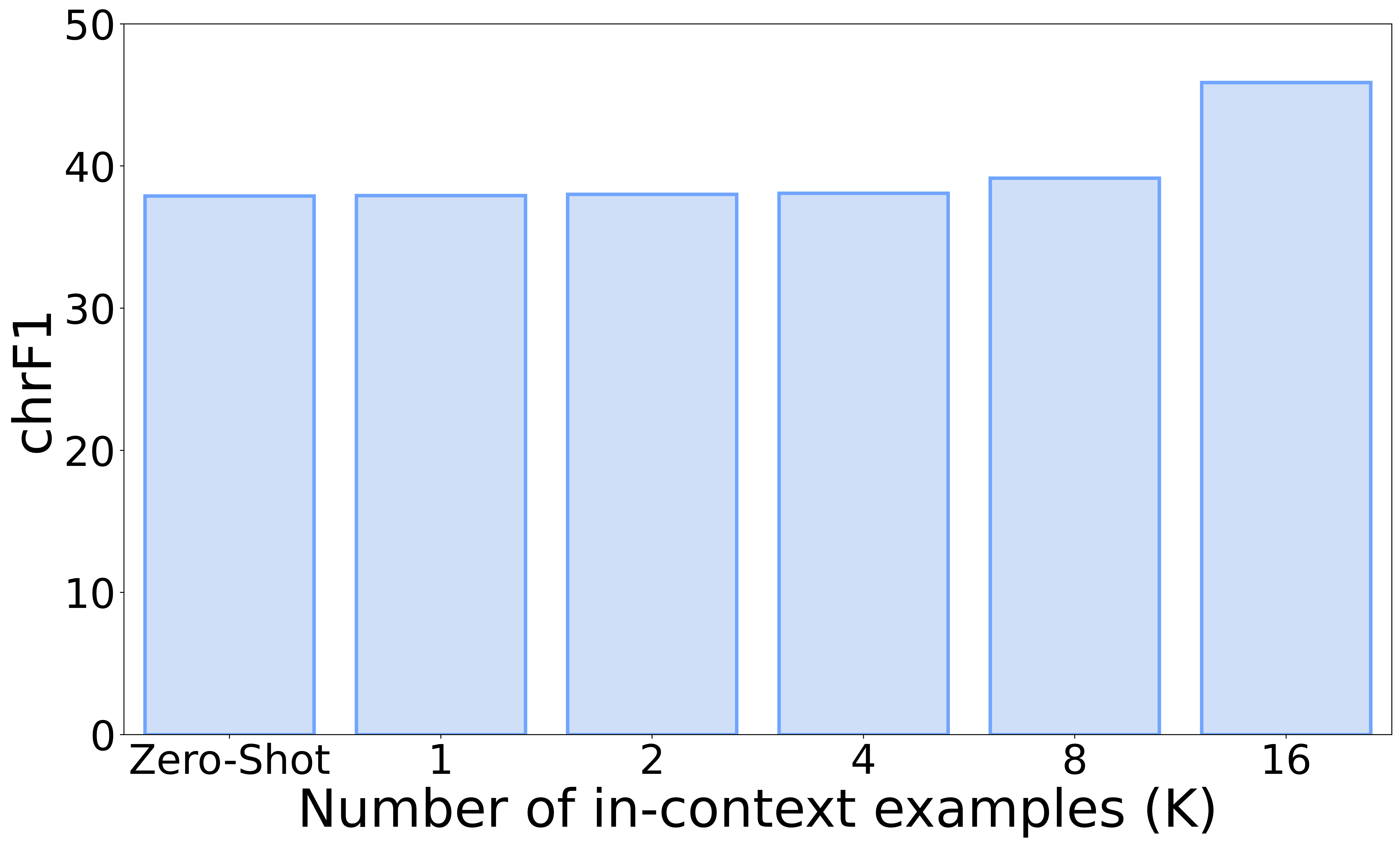}
    \caption{
    \small We plot the chrF-1 score for different values of $K\in\{1,2,4,8,16\}$, where K represents the number of in-context examples. The task for this experiment is translation from Rajasthani to English and LLM is LLAMA-3.1-8B.}
    \label{fig:ablations_icl}
\end{figure}

\paragraph{Ablation Study on the Number of In-Context Samples \(K\).} In Figure~\ref{fig:ablations_icl}, we present ablation results on the number of demonstrations, K, included in the prompt. Specifically, we evaluate the translation task from Rajasthani to English using chrF1 scores for outputs generated with varying \(K = \{1, 2, 4, 8, 16\}\). Due to computational constraints, we cap the maximum number of in-context examples at 16. Our results show that setting \(K = 16\) achieves the best performance.

\input{tables/xquad}
\input{tables/flores_gpt}

\section{Details on Diversity-induced finetuning}
\label{app:dpp}

We introduced DPP-based diversity finetuning in Section~\ref{subsubsec:dpp_finetune}. In this section, we provide additional details regarding the training procedure. We obtain the final retriever by fine-tuning $\rho^*$ on the merged dataset $\widetilde{\mathcal{D}} = \mathcal{D}^{\mathcal{T}} \cup \mathcal{D}^{\text{aux}}$. Specifically, for each sample $(\*x_i, \*y_i) \in \widetilde{\mathcal{D}}$, a subset of $\mathcal{E}_i$ in-context examples are retrieved from $\widetilde{\mathcal{D}}$. Out of the $\mathcal{E}_i$ subsets, a positive subset $E^{(+)}_i$ is selected using maximum a
posteriori (MAP) sampling~\citep{chen2018fastgreedymapinference} from the kernel matrix $\*Z$. The other $\mathcal{E}_i-1$ negative subsets ($E^{(-)}_i$) are selected using non-replacement random sampling, with no repeating examples in each subset. Based on these ground-truth sets, the retriever is fine-tuned using the following loss:

\begin{multline}
     \ell_{i} = \sum_{\left(E^{(+)}_i, E^{(-)}_i \right) \in \mathcal{E}_i} \max\{0, \log \det \left(\*Z_{E^{(-)}_i}\right) \\ - \log \det \left(\*Z_{E^{(+)}_i}\right) \},
\end{multline}
\begin{equation}
    \mathcal{L}_{\text{DPP}} = \frac{1}{\widetilde{N}} \sum_{i=1}^{\widetilde{N}} \ell_i,
\end{equation}

where $\widetilde{N}$ is the number of samples in $\widetilde{\mathcal{D}}$.

\section{Baseline Descriptions}
\label{app:baseline}

For a thorough and fair evaluation, we compare \name with the following baselines:

\begin{itemize}[noitemsep, leftmargin=*]
    \item Random: In-context examples are randomly selected from the training set without repetition. 
    \item BM25: A classical sparse retrieval method, BM25~\cite{robertson2009probabilistic}, is employed to rank and select the top-scoring examples as in-context samples. 
    \item Top-K: This baseline uses a dense retriever initialized with pre-trained multilingual BERT embeddings without any fine-tuning. The top-ranked examples are selected as in-context samples. 
    \item Diverse: We initialize the retriever with pre-trained BERT embeddings and apply Maximum a Posteriori (MAP) during inference to retrieve a diverse subset of examples.
    \item EPR~\citep{rubin2021learning}:  The multi-lingual BERT retriever is fine-tuned using the relevance loss and at the inference stage the top selected examples are selected for use as in-context samples.
    \item CEIL~\cite{ye2023compositional}: The retriever is fine-tuned using the DPP loss, incorporating both diversity and relevance. 
\end{itemize}

\section{Dataset Description}
\label{app:datasets}

\begin{enumerate}[noitemsep,nolistsep, leftmargin=*]
    \item Cross-Lingual Question-Answering (XorQA-In-XX): This task involves generating answers in non-English languages, given English evidence passages. Specifically, each example consists of a question in language XX, an English passage, and an answer in language XX, where the task is to generate the answer in the same language (XX) as the question. 

    \item Multilingual Question-Answering (XQuAD-IN): In this dataset, each example consists of a passage, question, and short answer in a source language (XX), and the task is to generate the answer in XX.

    \item Machine Translation (FLORES-IN-XX-En): For this dataset, the task is to translate a sentence from a source language (XX) to English. 

    \item Cross-Lingual Summarization (CrossSum-IN): Given a news article in a non-English language (XX), the task is to generate a summary of the article in the same language.
\end{enumerate}

\section{Prompts}
\label{app:prompts}

We present the prompts used for evaluation in Table~\ref{tab:prompts}
\input{tables/prompts}

\section{Algorithm}
\label{app:algorithm}

We elaborate the auxiliary dataset selection approach in Algorithm~\ref{algo:aux_select}.

\begin{algorithm*}[t]
\caption{Auxiliary Dataset Selection}
\label{algo:aux_select}
\begin{algorithmic}[1]

\State \textbf{Input:} High-resource language example bank   $\mathcal{D}^{\text{high}}=\{\mathcal{D}^{\mathcal{H}_1}, \cdots, \mathcal{D}^{\mathcal{H}_V}\}$; low-resource target language examples $\mathcal{D}^{\mathcal{T}}=\left\{ (\*x_{i}, \*y_{i}) \right\}_{i=1}^{N}$; pre-trained multi-lingual BERT encoder $\phi$; Threshold $\delta$.

\State $e_{\mathcal{T}} \gets \cfrac{1}{N}\sum_{i=1}^{N}\phi((\*x_{i}, \*y_{i}))$ \Comment{Compute average of BERT embedding for low-resource samples}

\State $\text{sim} \gets \emptyset$

\For{each $\mathcal{D}^h \in \mathcal{D}^{\text{high}}$}
    \State $e_{h} \gets \cfrac{1}{N_h}\sum_{j=1}^{N_h}\phi((\*x_{jh}, \*y_{jh}))$ \Comment{Compute mean BERT embedding for each auxiliary dataset}
    \State $\text{sim}_{h} \gets \cfrac{e_{h}^{\top}e_{\mathcal{T}}}{|e_{h}||e_{\mathcal{T}}|}$

    \State $\text{sim} \gets \text{sim} \cup \text{sim}_{h}$

\EndFor

\State $\mathcal{D}^{\text{aux}} \gets \{\mathcal{D}^h \in \mathcal{D}^{\text{high}} \mid \text{sim}_h \geq \text{percentile}(\text{sim}, \delta)\}$

\State \Return $\mathcal{D}^{\text{aux}}$ 
\end{algorithmic}
\end{algorithm*}

%% file: tables/xquad.tex
\newcolumntype{?}{!{\vrule width 1pt}}
\newcolumntype{a}{>{\columncolor{myblue}}c}
\begin{table}[!t]
      \centering
        \resizebox{\columnwidth}{!}{%
        \begin{tabular}{cccccc}
        \toprule
       \multirow{2}{1.75cm}{\centering Aux. Data Used} & \multirow{2}{1.75cm}{\centering Finetuning-Based} & \multirow{2}{*}{Methods} & \multicolumn{3}{c}{Odia}  \\ 
       \cmidrule{4-6}

       & & & QW-2-7B & LM-3.1-8B & QW-2.5-7B \\

       \midrule
       \xmark & \xmark &  Zero-shot & 12.67 & 13.08 & 21.76  \\
      \xmark & \xmark & Random & 25.55 & 33.60 & 34.18  \\
       \xmark & \xmark & BM25 & 31.10 & 32.46 & 34.89  \\
        \xmark & \xmark & Top-K & 28.90 & 29.30 & 33.20 \\
       \xmark & \xmark & Diverse & 28.37 & 31.70 & 33.98  \\
        \xmark & \cmark & EPR & 32.36 & 34.91 & 36.51  \\
      \xmark & \cmark & CEIL & 33.11 & 36.70 & 36.37  \\
       \midrule
        \cmark & \cmark & EPR & 32.79 & 37.02 & 37.18 \\
        \cmark & \cmark & CEIL & 32.40 & 39.88 & 36.28 \\
     \rowcolor{myblue} \cmark & \cmark & \name~(Ours) & \textbf{35.92} & \textbf{46.87} & \textbf{43.65}  \\
     & & Absolute Gain $(\Delta)$ & +2.81 & +6.59 & +6.47 \\
      \bottomrule
    \end{tabular}%
        }
 \caption{\small \textbf{Evaluation on Xquad-In.} We report Token-F1 results for QA performance from Odia to English. The evaluation includes three LLMs: Qwen-2-7B (QW-2-7B), Qwen-2.5-7B (QW-2.5-7B), and LLAMA-3.1-8B (LM-3.1-8B).}
\label{tab:xquad}
\end{table}

%% file: tables/flores_gpt.tex
\newcolumntype{?}{!{\vrule width 1pt}}
\newcolumntype{a}{>{\columncolor{myblue}}c}
\begin{table}[!t]
      \centering
       
        \resizebox{\columnwidth}{!}{%
        \begin{tabular}{ccccc?cc}
        \toprule
       \multirow{2}{1.75cm}{\centering Aux. Data Used} & \multirow{2}{1.75cm}{\centering Finetuning-Based} & \multirow{2}{*}{Methods} & \multicolumn{2}{c?}{sat} & \multicolumn{2}{c}{mni} \\
       \cmidrule{4-7}

       & & & GPT-3.5 & GPT-4 & GPT-3.5 & GPT-4\\

       \midrule
       \xmark & \xmark &  Zero-shot &  15.57 & 15.98 & 22.21 & 27.90 \\
       \xmark & \xmark & Random &  14.34 & 15.21 & 20.19 & 26.98\\
       \xmark & \xmark & BM25 &  16.52 & 16.77 & 25.61 & 29.05\\
       \xmark & \xmark & Top-K & 16.63 & 17.12 & 25.80 & 30.21 \\
      \xmark & \xmark & Diverse & 16.27 & 16.59 & 24.85 & 29.01\\
      \xmark & \cmark & EPR & 18.35 & 18.71 & 26.93 & 31.26 \\
      \xmark & \cmark & CEIL & 18.62 & 19.13 & 27.31 & 32.09 \\
       
       \midrule
      
       \cmark & \cmark & EPR & 18.27 & 18.58 & 26.49 & 30.88 \\
      \cmark & \cmark & CEIL & 18.61 & 19.02 & 27.03 & 31.85\\
     \rowcolor{myblue} \cmark & \cmark & \name (Ours) & \textbf{19.33} & \textbf{20.98} & \textbf{28.38} & \textbf{34.35} \\
    
      \bottomrule
    \end{tabular}%
        }
 \caption{\small \textbf{Evaluation on Proprietary LLMs.} We report chrF1 results for translation performance from Santali (sat) and Manipuri (mni) to English on GPT-3.5/GPT-4. For this evaluation, we used LLAMA-3.1-8B as the scorer LLM.}
\label{tab:flores_close}
\end{table}

%% file: tables/prompts.tex
\newcolumntype{?}{!{\vrule width 1pt}}
\newcolumntype{a}{>{\columncolor{myblue}}c}

\begin{table*}[!t]
\centering
\resizebox{\textwidth}{!}{%
\begin{tabular}{c?l}
\toprule
\textbf{Dataset} & \textbf{Prompt}\\
\midrule

\multirow{4}{*}{CrossSum-In} & \textcolor{red}{[Context]} \\
& Summarize the article in \textcolor{teal}{[Target Language Name]} language. \\
& Summarize the following article: \textcolor{blue}{[Article in English]} \\
& Summary: \\
\midrule
\multirow{5}{*}{XorQA-In-XX} & \textcolor{red}{[Context]} \\
& Generate an answer in \textcolor{teal}{[Target Language Name]} language for the question based on the given passage. \\
& \textcolor{blue}{[Passage in English]} \\
& Question: \textcolor{blue}{[Question in Target Language]} \\
& Answer: \\
\midrule
\multirow{4}{*}{Flores-In} & \textcolor{red}{[Context]} \\
& Translate the following sentence to English. \\
& Input: \textcolor{blue}{[Sentence in Target Language]} \\
& Output: \\
\midrule
\multirow{4}{*}{XQuAD-In} & \textcolor{red}{[Context]} \\
& Generate an answer for the next question in \textcolor{teal}{[Target Language Name]} language. \\
& \textcolor{blue}{[Passage in Target Language ]} \\
& Question: \textcolor{blue}{[Question in Target Language]} \\
& Answer: \\

\bottomrule
\end{tabular}%
}
\caption{Prompt template used for various datasets.}
\label{tab:prompts}
\end{table*}